\pdfoutput=1
\documentclass{article}

\usepackage{arxiv}

\usepackage[utf8]{inputenc} 
\usepackage[T1]{fontenc}    
\usepackage{hyperref}       
\usepackage{url}            
\usepackage{booktabs}       
\usepackage{amssymb,amsfonts}       
\usepackage{nicefrac}       
\usepackage{microtype}      
\usepackage{lipsum}
\usepackage{graphicx}
\usepackage{multirow}
\usepackage{algorithmic}
\usepackage{algorithm}
\graphicspath{ {./} }
\usepackage{subfigure}

\title{CIDMP: Completely Interpretable Detection of Malaria Parasite in Red Blood Cells using Lower-dimensional Feature Space}

\author{
Anik Khan\\
  University of Memphis\\
  Tennessee, USA \\
  \texttt{akhan9@memphis.edu} \\
  \And
 Kishor Datta Gupta \\
  University of Memphis\\
  Tennessee, USA \\
  \texttt{kgupta1@memphis.edu} \\
  \And
Deepak Venugopal \\
   University of Memphis\\
  Tennessee, USA \\
  \texttt{dvngopal@memphis.edu} \\
   \And
Nirman Kumar \\
   University of Memphis\\
  Tennessee, USA \\
  \texttt{nkumar8@memphis.edu} \\
}

\begin{document}
\maketitle
\begin{abstract}

Predicting if red blood cells (RBC) are infected with the malaria parasite is an important problem in Pathology. Recently, supervised machine learning approaches have been used for this problem, and they have had reasonable success. In particular, state-of-the-art methods such as Convolutional Neural Networks automatically extract increasingly complex feature hierarchies from the image pixels. While such generalized automatic feature extraction methods have significantly reduced the burden of feature engineering in many domains, for niche tasks such as the one we consider in this paper, they result in two major problems. First, they use a very large number of features (that may or may not be relevant) and therefore training such models is computationally expensive. Further, more importantly, the large feature-space makes it very hard to interpret which features are truly important for predictions. Thus, a criticism of such methods is that learning algorithms pose as opaque blackboxes to its users, in this case medical experts. The recommendation of such algorithms can be understood easily, but the reason for their recommendation is not clear. This is the problem of non-interpretability of the model, and the best-performing algorithms are usually the least interpretable.  To  address these issues, in this paper, we propose an approach to extract a very small number of aggregated features that are easy to interpret and compute, and empirically show that we obtain high prediction accuracy even with a significantly reduced feature-space.
\end{abstract}


\section{Introduction}
\label{sec1}
Malaria is a serious disease caused by a parasite and transmitted by mosquito bites. It is most severe on pregnant women, immune compromised patients, and children under the age of 5. It kills more than 3000 children a day in Africa \cite{unicef_2003}. In 2018, the number of malaria deaths reached around 405 000 \cite{world_health_organization}. While there are treatments available for Malaria, early detection and 
intervention is important for successful recovery. As such detection of the disease is an important problem. Unfortunately, the symptoms 
of Malaria are not very distinguished, even though they may be acute. A blood test followed by a pathologist's inspection of the sample
are important for diagnosis. Artificial Intelligence (AI) that can aid the pathologist in this diagnosis can be a game changer in 
saving pathologist time.

While AI research has seen many peaks and valleys in its six decades journey \cite{holland1992adaptation,russell1995artificial}, the field has recently gained immense attention, mostly due to the successful implementation of  Machine Learning (ML) to solve many problems (such as automatic speech recognition, hand writing recognition, weather forecasting, etc.) typically belonging to the AI domain. On the other hand, AI has fueled the need for ML to develop algorithms that can learn autonomously from data \cite{shahriari2015taking}.
To achieve a state of usable intelligence we need an automatic workflow that will learn from prior data,
will extract information, will generalize, and will expose the hidden informative factors of the data \cite{bengio2013representation}.
Even with such a workflow, a remaining challenge is to make sense of the data in the context of an application domain. While data-quality assurance and feature extraction are significant steps \cite{girshick2014rich} in ML algorithm design, the full effectiveness of success is limited by the algorithm's inabilities to explain its results to human experts. This is a prime concern in applications of AI to the medical domain. Indeed, the results of an
AI algorithm need to be ``explainable'', i.e., an expert should be
able to understand the \emph{raison d'\^etre} for the results \cite{samek2017explainable}. 

In particular, in the medical domain, interpretability of an ML method is crucial since it involves making decisions that have far-reaching consequences\cite{samek2017explainable}. Therefore, it is critical that ML applications in healthcare are transparent and trustworthy\cite{samek2017explainable}. However, there is typically a tradeoff between performance (predictive accuracy) and explainability. Often the best-performing methods (such as deep learning) are the least explainable, and the ones providing a clear explanation (like decision trees) are less accurate \cite{bologna2017characterization}.
In general, the performance of ML algorithms relies heavily upon feature representation; hence most of the effort traditionally goes into pre-processing and feature engineering. Over the last several years though, approaches such as deep learning have aimed towards automatically extracting features from the data devoid of any domain expertise. While this is extremely useful, it has certain drawbacks. Specifically, the feature-space that typical neural networks operate in, is extremely large. For example,  famous CNN architectures such as Resnet\cite{targ2016resnet}, Alexnet\cite{krizhevsky2012imagenet}, etc. use large number of parameters and produce extremely large feature-spaces. While this improves the performance of ML on very challenging tasks, the large feature-space is problematic since it is extremely hard to separate the discriminating information from the overall information. Further, learning such models require vast amounts of data and computational resources \cite{hassannejad2016food}. In this paper we consider the problem of predicting malaria parasites from RBC images. We develop a very small number of aggregated features and show that even with a small feature-space we obtain high prediction accuracy. 


Our objective in this paper is to study the efficacy of aggregated features from multispectral RGB features to distinguish malaria parasite infected RBC from non-infected RBC. Thus, two classes are considered: ``Infected'' and ``Uninfected''. We compute aggregated interpretable features, that help to reduce the problem dimensionality significantly. As our features are interpretable, this beats any deep learning model in explainability, interpretablility and trust. Then using these aggregated features, we trained a prediction model.  To achieve this goal, the Random Forests classifier is used. This algorithm is well suited for classification and can be trained with large datasets as the number of features is small. In addition, it measures feature importance so that the relative importance of a feature can be examined for the different classes we consider. In this work, we have studied feature importance by generating histograms for both classes. As the feature space is substantially lower-dimensional than the input image space, the feature importance is not interpretable enough for medical experts. Therefore, we present an explainer for the model in the input space.

\textbf{Challenges. } An essential criterion for explanations is interpretability \cite{ribeiro2016should}. 
One reason why feature importance is not an effective explanation is because it makes an unrealistic expectation of the medical expert:
to perceive directly how a feature is related to the image.
To be effective, an explainer has to provide a qualitative understanding between the input image and the prediction of the model. 
Thus, a linear \cite{kononenko2010efficient}, a gradient vector \cite{baehrens2010explain} or an additive model \cite{caruana2015intelligible} may or may not be interpretable. For example, if a single specific aggregated feature is important to a prediction, it is not reasonable to expect any medical expert to understand why the prediction was delivered.  Therefore, explanations should be simple to understand, which may not be true for most of the aggregated features used by the model. In addition, we note that the
idea of interpretability also depends on the target audience. So, if one medical expert is skilled at a specific image domain (such as RBC images), it may not be effective to provide 2D frequency spectrum image or explanation in texts/audio.

Another challenge is that the explainer has to be effective for specific inputs locally, and not just globally, in a sensitive field like a medical diagnostics. Local model interpretation is a set of procedures attempting to understand individual predictions of machine learning models. It has less value for medical experts to make a decision based on a model's prediction if explainer provides global interpretation. Global model interpretation is a set of values (mostly statistical measures of input variables or feature values) attempting to understand how predictions are made by a machine learning model.

In this paper, we propose a new framework called CIDMP that is sufficiently robust and explainable for use in malaria-infected RBC diagnosis. We computed explainable features from RBC images that are mathematically and informatically sound in malaria parasite detection. We adopt an explainable modeling approach so as to obtain better interpretability and generalizability.

\begin{figure*}
\centering
\subfigure{
\includegraphics[width=2.5in]{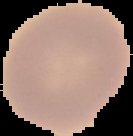}
	\label{rbc1}%
}
\subfigure{
\includegraphics[width=2.5in]{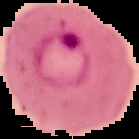}
	\label{rbc2}%
}
\caption{Uninfected RBC (left) and an infected RBC (right)}
	\label{rbc}%
\end{figure*}

\section{Related Work}
\label{sec4}
This section summarizes the attempts made by researchers to diagnose malarial parasites using digital image processing (DIP) algorithms. Some of these methods are complex and need manual supervision. Other methods using neural networks are black boxes.
A $k$-nearest neighbor (KNN) classifier was proposed for diagnosis and screening of malaria \cite{tek2009computer}. It also considers color-based and shape-based features for detecting parasites and non-parasites.

SVM classifiers have been used to  classify
parasite-infected areas using color, size, and textural features\cite{linder2014malaria}.
The paper \cite{savkare2011automatic} used Otsu thresholding and watershed transforms for segmentation of blood cells and then applied SVM to detect parasites using color and statistical features.

RGB, HSI and C-Y color spaces were studied in \cite{abdul2013colour} to detect malaria parasites. The paper discusses the mathematical model for classification of RBC as infected parasite and non-parasite. The method for counting RBC is also discussed. Comparative analysis of different classifiers for malaria detection is presented in \cite{khot2015optimal}. In \cite{tsai2015blood} a technique is proposed to automate segmentation for malaria parasite and infected erythrocytes detection. The automatic counting of the number of malaria parasites using conventional image processing algorithms such as histogram equalization, thresholding, morphological operations, and connected components analysis is demonstrated in \cite{arco2015digital}.

Another line of research is to apply machine learning methods to medical image analysis. The aim is to get
meaningful feature representations that are significant to achieve desired results. The majority of computer-aided diagnosis (CADx) software uses machine learning techniques with deterministic features for decision-making \cite{ross2006automated,das2013machine,poostchi2018image}. However, the process requires expertise in examining the variability in size, ratio, background, angle, region of interest (ROI), and signal to noise ratio (SNR) of the images. To overcome the challenges of devising deterministic features that capture variations in the underlying data, Deep Learning is used with notable success \cite{lecun2015deep}. Deep learning models use
a cascade of layers to self-discover the relation between input raw data and prediction. Higher-level features are extracted from lower-level features to aid in learning complex decision-making functions, resulting in end-to-end feature extraction and classification \cite{schmidhuber2015deep}. Unlike kernel-based algorithms like Support Vector Machines (SVMs), deep learning models usually exhibit enhanced performance with an increase in data size and computational resources \cite{srivastava2014dropout}.

In \cite{rajaraman2018pre} a CNN (Convolutional Neural Network) based model is proposed. The model has three convolutional layers and two fully connected layers to classify malaria parasite from RBC image.
Our work uses domain knowledge to analyze and compute features. We harness the shape information of the malaria parasite and demonstrate satisfactory interpretability both in feature space and model prediction, with high accuracy.

\section{Proposed Method}

\begin{figure*}[!htb]
\center{\includegraphics[scale=0.3]{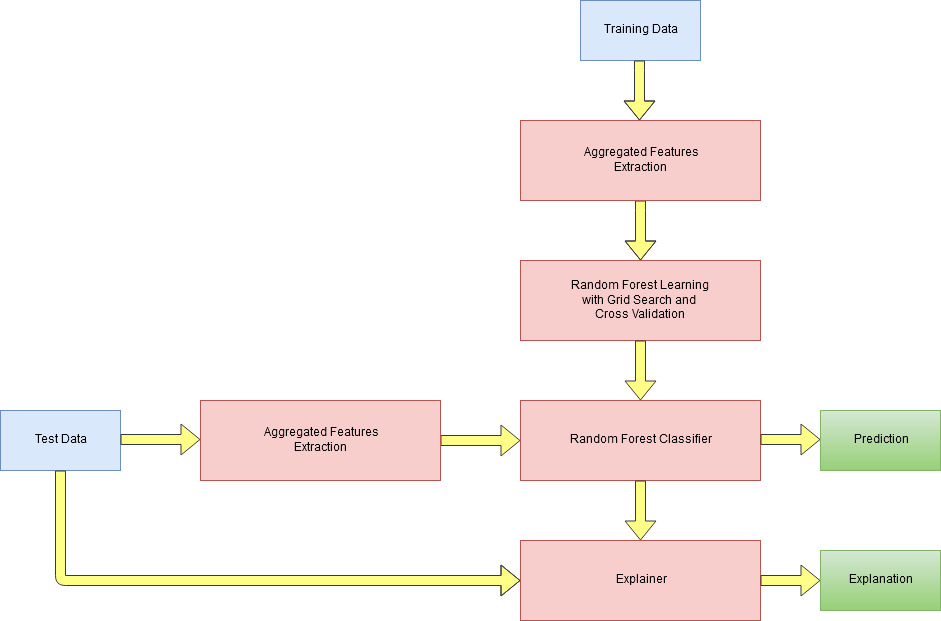}}
\caption{Summary of workflow of proposed method}
	\label{flowchart}%
\end{figure*}

The problem we want to solve is to detect the presence of malaria parasite in Red Blood Cell image. \ref{rbc} shows an uninfected RBC (left) and an infected RBC (right). Our method is based on the following observation: The malaria infected RBC has a ring shaped entity whereas the uninfected RBC doesn't contain such a ring shaped entity. Thus, our proposed method is to detect the ring shaped entity as shown in \ref{flowchart}. At first we extract aggregated features from RBC images. Then we train a random forest classifier with extracted features. After proper training, we use this trained classifier to classify malaria infected RBC and provide an interpretation of the decision. For sanity and trust reasons, we need interpretable features rather than features accumulated in several hidden layers inside a deep learning model. 

Now we propose the features to capture the ring shaped structure in RBC.

\subsection{Feature Extraction}
\label{subsec1}

\subsubsection{Aggregated Laplacian coefficient}
\label{subsubsec1}
Laplacian of an image provides significant information in image classification field \cite{gao2010local,gomez2008semisupervised,tu2011laplacian}.
The Laplacian is a 2-D isotropic proportion of the second spatial derivative of an image \cite{haralick1992computer}. The Laplacian of an image features the fast changes in pixel intensity values and is mostly utilized for edge detection problems. The Laplacian is generally used on gray-level images or single-channel images. Laplacians of the images were calculated separately for each channel (red, blue and green). Aggregated laplacian coefficient were calculated by summation of each laplacian.
Let $I(x,y)$ denotes a pixel in RBG image at $(x,y)$ position. $I_r(x,y)$, $I_g(x,y)$ and $I_b(x,y)$ are red, green and blue channels' pixel intensity of $I$ at $(x,y)$ position. We denote the Laplacian at position $(x,y)$ from $R$ channel by $L_r(x,y)$, and similarly for blue and green.

\begin{equation}
L_r(x,y) = \frac{\partial^2 I_r}{\partial x^2} + \frac{\partial^2 I_r}{\partial y^2},
\end{equation}

where the partial derivatives are computed using finite difference methods. 
We denote $I_{alc}(r)$ as aggregated Laplacian coefficient for $R$ channel.

\begin{equation}
    I_{alc}(R) = \sum_x \sum_y L_r(x,y)
\end{equation}

Similarly, We compute $I_{alc}(G)$ and $I_{alc}(B)$ from 
$ L_g(x,y)$ and $ L_b(x,y)$ respectively.

\begin{figure}%
\centering
\subfigure{%
\includegraphics[width=3.0in, height=1.5in]{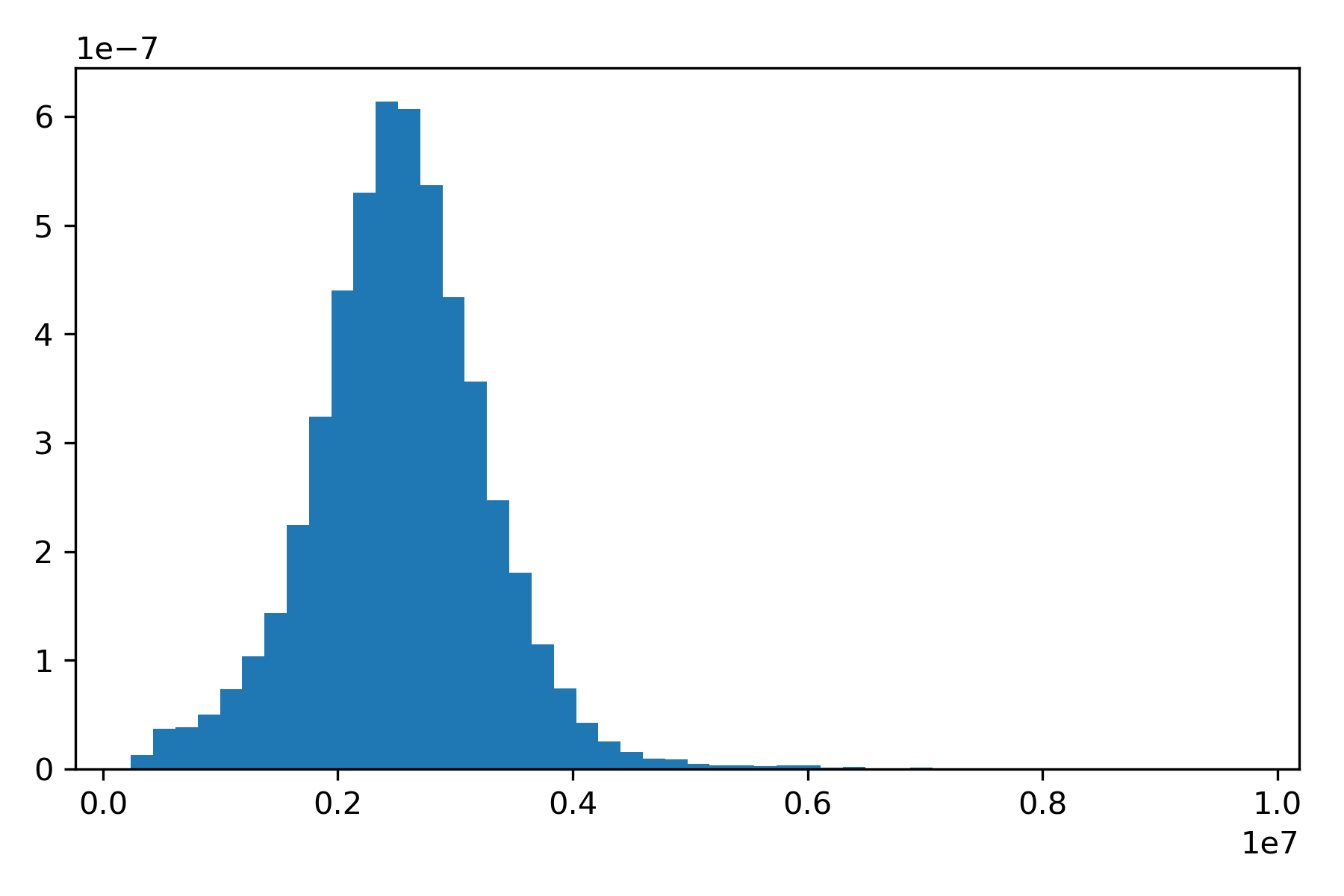}
\label{laplacianRed1}%
}%
\subfigure{%
\includegraphics[width=3.0in,height=1.5in]{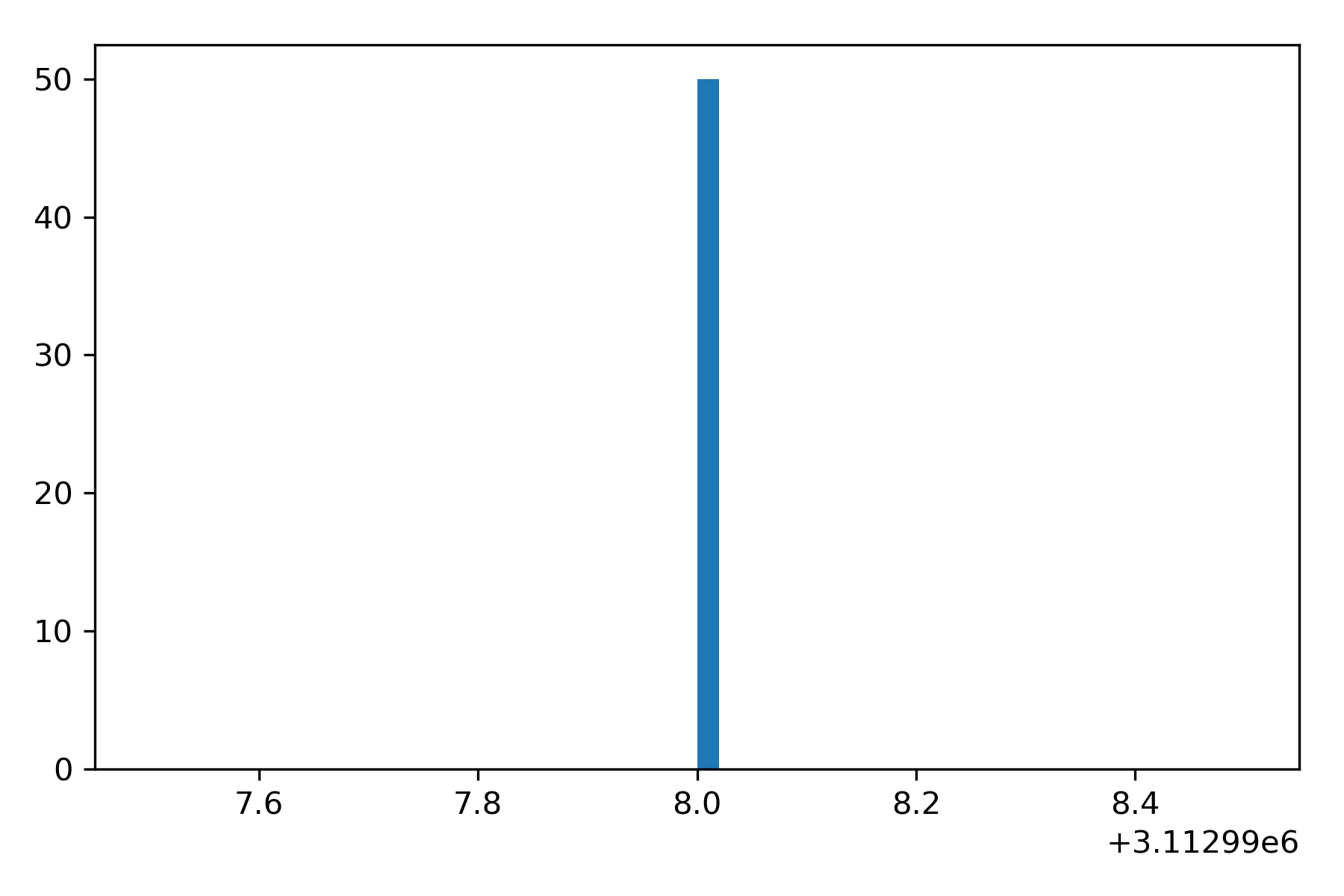}
	\label{laplacianRed2}%
}%
\caption{Histogram of aggregated Laplacian coefficient on Red channel of infected (left) and uninfected(right) RBC}
	\label{laplacianRed}
\end{figure}

\begin{figure*}%
\centering
\subfigure{%
\includegraphics[width=3.0in,height=1.5in]{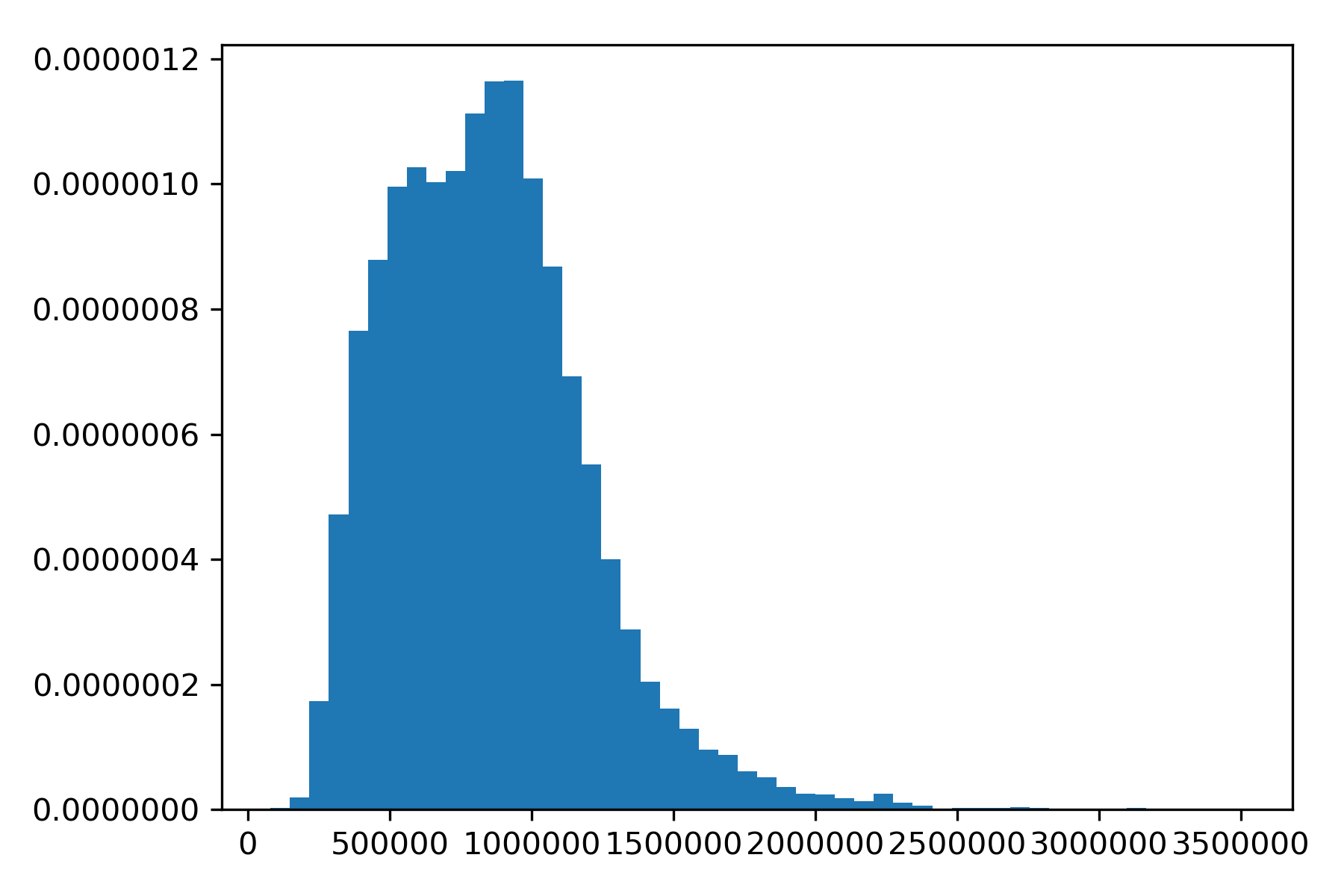}
	\label{laplacianBlue1}%
}%
\subfigure{%
\includegraphics[width=3.0in,height=1.5in]{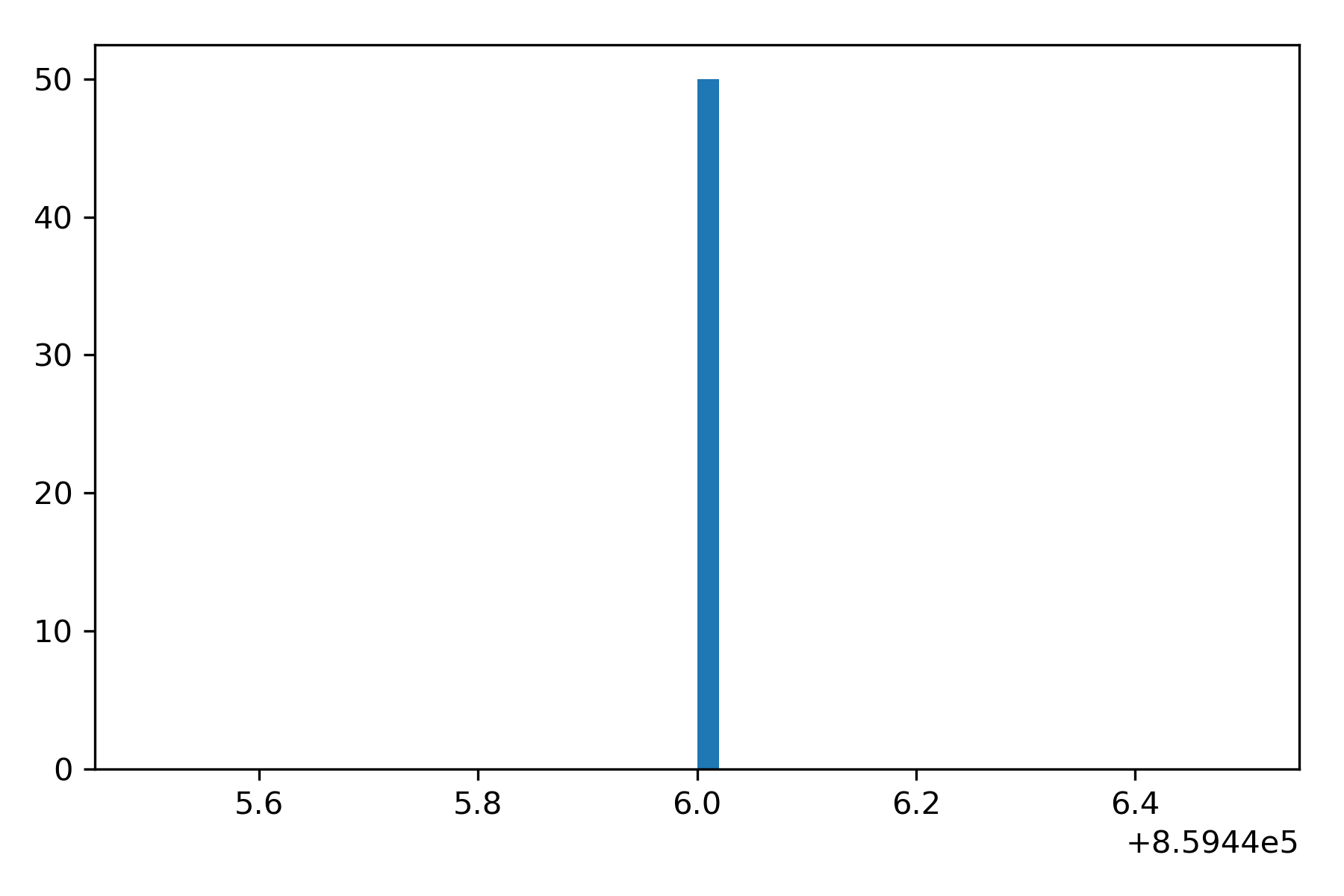}
	\label{laplacianBlue2}%
}%
\caption{Histogram of aggregated Laplacian coefficient on Blue channel of infected (left) and uninfected(right) RBC}
	\label{laplacianBlue}
\end{figure*}

\begin{figure*}%
\centering
\subfigure{%
\includegraphics[width=3.0in,height=1.5in]{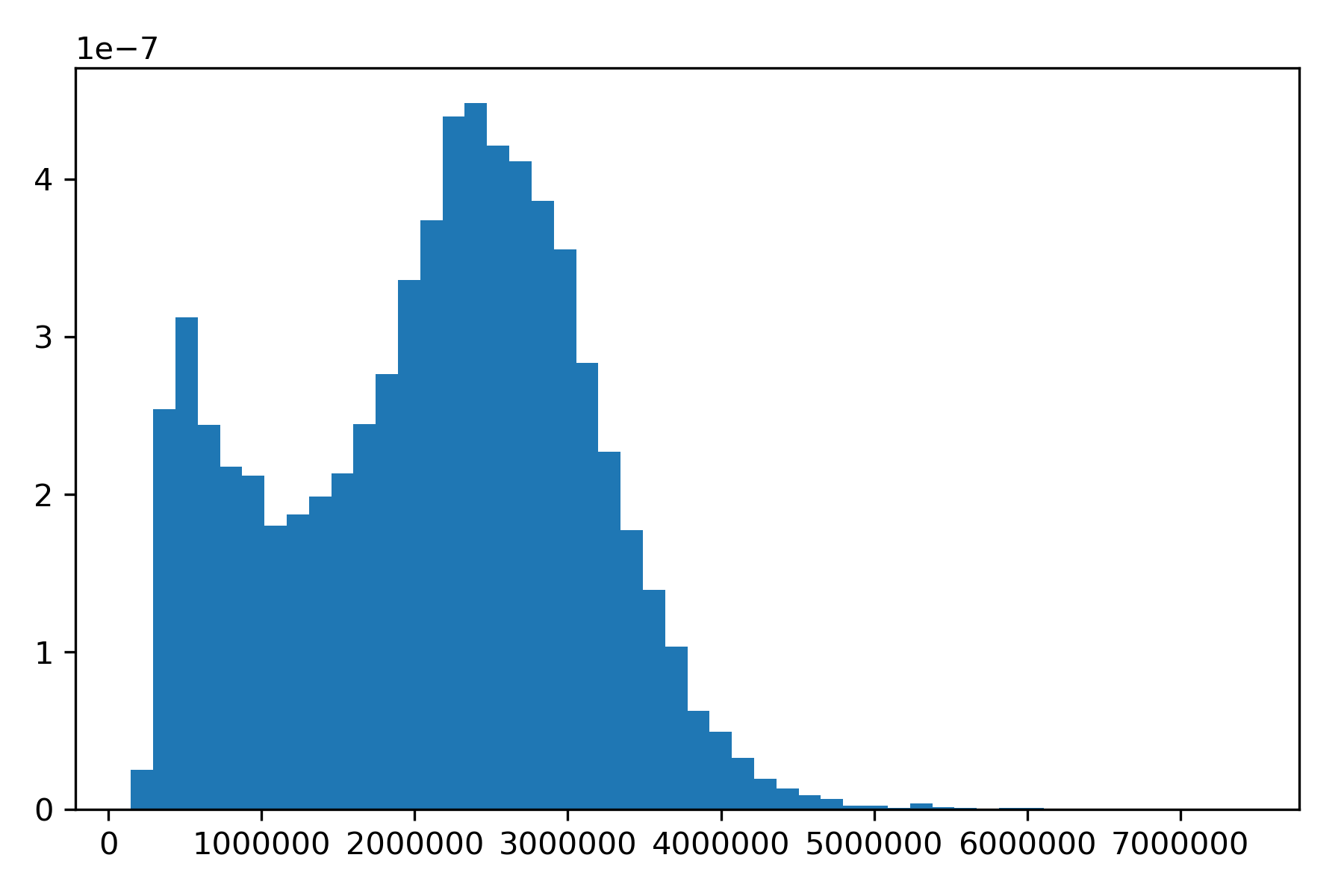}
	\label{laplacianGreen1}%
}%
\subfigure{%
\includegraphics[width=3.0in,height=1.5in]{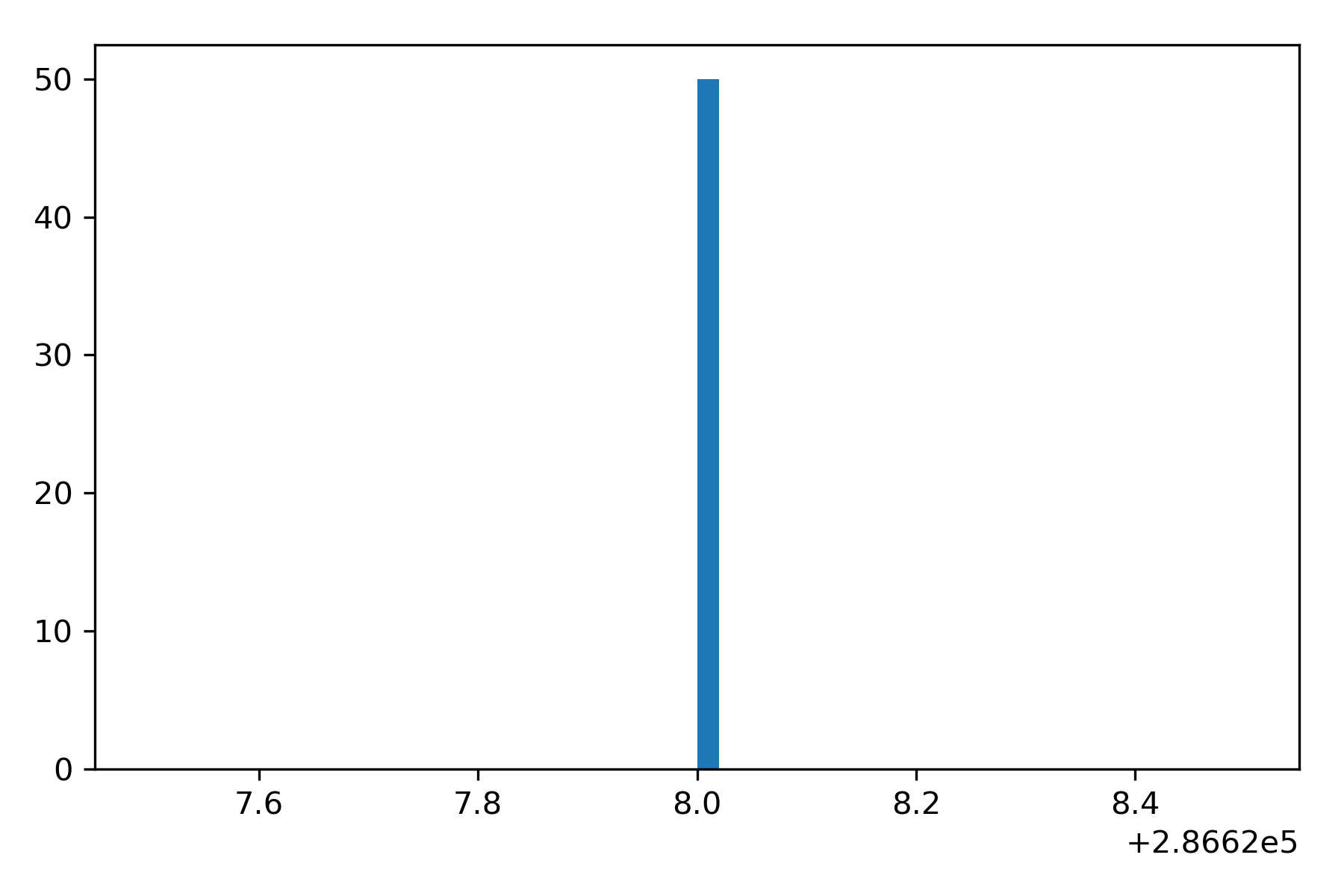}
	\label{laplacianGreen2}%
}%
\caption{Histogram of aggregated Laplacian coefficient on Green channel of infected (left) and uninfected(right) RBC}
\label{laplacianGreen}
\end{figure*}

Histogram showed (see \ref{laplacianRed}, \ref{laplacianBlue}, and \ref{laplacianGreen}) significant difference present in the distribution of aggregated laplacian coefficient between infected and uninfected RBC images. Therefore, the aggregated laplacian coefficient promises to be an important feature for distinguishing infected and uninfected images.

\subsubsection{Inner Ring length}
\label{subsubsec2}
As malaria parasite is endothelial to RBC surface, the ring shape of the parasite is visible in RBC images. This length conceivably meddles with the deformation of the membrane of RBC \cite{hosseini2012malaria}. So, this inner ring detection is a challenging and important step for building an interpretable classifier.

To compute the length of the inner ring we first filter the image with an edge detection algorithm, and then remove the outer boundary of RBC image.
\begin{figure*}%
\centering
\subfigure{%
\includegraphics[width=3.5in]{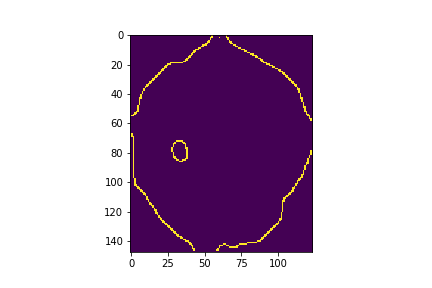}}%
\subfigure{%
\includegraphics[width=3.5in]{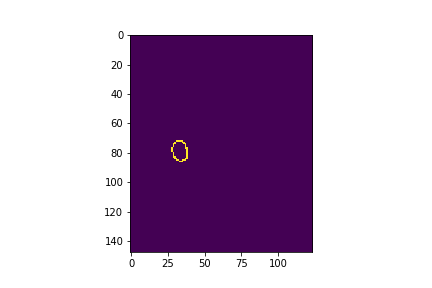}}%
\caption{Output of Canny algorithm (left) and result after boundary removal (right)}
	\label{canny}
\end{figure*}
We have used the Canny edge detection algorithm \cite{canny1986computational} to detect the edges of an image. The Canny edge detection algorithm has several advantages over others (i.e. Robert, Sobel, Prewitt ) \cite{maini2009study, nisha2015comparative}. Firstly,  Canny operation is not very susceptible to noise. So, it is less probable that we will extract an inner ring induced from noise.
Secondly, since we are trying to calculate the length of the inner ring, our ring extractor algorithm should be able to produce smooth and thin edges. Canny algorithm does it better than Sobel \cite{maini2009study}. \ref{canny} shows the output of Canny algorithm (left) and the inner ring extraction after boundary removal (right).

After getting the edges from the image, we remove the outer boundary. This will give us an image with only the inner ring in it. Then we count the number of pixels containing edges to get an estimation of the length.

\subsection{Model}
The random forest, also known as the decision forest model, is an ensemble learning strategy for both classification and regression problems. It works by building a large number of decision trees at preparation time, and yielding the class that is the method of the classes (grouping) or mean expectation (relapse) of the individual trees \cite{ho1995random,barandiaran1998random}. The model mitigates the issue of overfitting in the decision tree model.
We selected Random Forest as our classifier for several reasons. Firstly, each decision tree has a large variance with low bias. However, in the random forest, all the trees vote independently, so the variance gets averaged. So random forest provides a low bias and moderate variance model. Secondly, Random forest examines outliers by essentially binning them. Thirdly, Random forest  works well with non-linear features. Finally, Random forest works well for image classification problems \cite{millard2015importance,bosch2007image,horning2010random}.

\subsection{Explainer}
By ``explaining a prediction'', we mean visual artifacts that present qualitative understanding of the relationship between the pixels of RBC images and the model's prediction. 
In \cite{ribeiro2016should} the process of explaining individual predictions is presented. It is clear that a hematologist, or even medical doctors, are much better positioned to make a decision with the help of a model if they have access to the explanations.
For RBC images, we present the annotation of pixels of a single image that convinced the model to classify that image as infected or uninfected.

In this work, we use LIME\cite{ribeiro2016should}, as an explanation technique that explains the predictions of any classifier in an interpretable way. The main idea is to train an interpretable model locally around the decision boundary of the prediction. Using LIME, we explain our model by exhibiting representative individual prediction. The main purpose of using LIME is to understand the behavior of the Random forest model by perturbing the input image and perceive how the predictions change. 

This happens to be a benefit in terms of interpretability, because we can perturb the input by changing components that make sense to medical experts, even though the model may be using more convoluted components as features (such as aggregated laplacian coefficients).
LIME generates an explanation by approximating the underlying Random forest model by an interpretable one (such as a linear model with only a few non-zero coefficients), learned on perturbations of the input image instance (hiding parts of the image). The key intuition behind LIME is that it is much easier to approximate a complex model by a simple model locally (in the neighborhood of the prediction), as opposed to trying to approximate a model globally. This is done by weighting the perturbed images by their similarity to the input image instance that needs an explanation. If the perturbed image instance has decent similarity with the input image instance, then LIME infer the pixels of the perturbed image significant for prediction (infected or uninfected) concerning the Random forest model.

\section{Experimental Results and Analysis}
Level-set based method was used to detect and segment the RBCs. Multi-scale Laplacian of Gaussian filter was applied to detect centroids of individual RBCs. The generated markers were used for cell segmentation inside a level set active contour framework to restrict the dynamic shape to the cell boundary. Several morphological operation were performed in post-processing to remove  the falsely detected molecules such as staining artifacts, using average cell size. White blood cells (WBCs) were removed using a one-to-one correspondence based on cell ground-truth annotations. This seems to be reflective of the images that are presented to medical experts under usual circumstances \cite{datasource}.

The segmented cells from the thin blood smear slide images for the infected and uninfected classes are downloaded from \cite{datasource}. The dataset contains a total of 27,558 cell images with equal instances of infected and uninfected cells.

We chose 80/20 training-test split. We implemented a gridsearch mechanism with 5-fold cross validation to get the best random forest predictor. In our run, the best random forest predictor had 25 estimators, \ref{prec-recall}, and \ref{roc} show the characteristics of our classifier. \ref{comparative-perf1} shows that random forests gives better performance in terms of True Positives, True Negatives and False Negative. It is noteworthy that random forest gives around twice better accuracy in terms of False Negative which is extremely important in medical diagnostics. If we implement a system with low performance with False Negative, there is higher probability that the diseases will go undetected. \ref{comparative-perf2} shows Random Forest performs better than other training models in terms of precision, recall and F1-score.

\begin{center}
\begin{table}
\begin{center}
\begin{tabular}{ |c|c|c|c|c| } 
 \hline
 Learning Algorithm & TN & FP & FN & TP \\ 
 \hline
 
 Logistic Regression & 2361 & 394 & 671 & 2085 \\
 Decision Tree & 2174 & 581 & 675 & 2081 \\
 \bf{Random Forest} & \bf{2258} & \bf{497} & \bf{397} & \bf{2359}  \\

 \hline
\end{tabular}
\caption{ Comparative performance: TN, FP, FN, TP}
\end{center}
\label{comparative-perf1}%
\end{table}
\end{center}

\begin{center}
\begin{table}
\begin{center}
\begin{tabular}{ |c|c|c|c|} 
 \hline
 Learning Algorithm &  Precision & Recall & F1 Score\\ 
 \hline
 
 Logistic Regression &  0.84 & 0.75 & 0.79 \\
 Decision Tree &  0.78 & 0.75 & 0.76\\
 \bf{Random Forest} &  \bf{0.82} & \bf{0.86} & \bf{0.84} \\

 \hline
\end{tabular}
\end{center}
\caption{ Comparative performance: Precision, recall, F1-score}
\end{table}
\label{comparative-perf2}%
\end{center}





\begin{figure}[!htb]
\center{\includegraphics[scale=0.4]{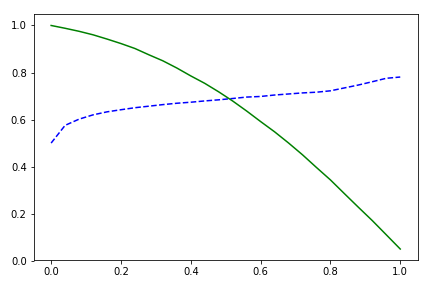}}
\caption{Precision$-$Recall vs. Threshold curve, Precision is indicated by dotted blue and recall is indicated by solid green}
	\label{threshold}%
\end{figure}

\begin{figure}[!htb]
\center{\includegraphics[scale=0.4]{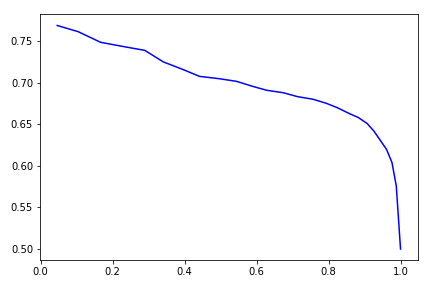}}
\caption{Precision vs. Recall Curve}
\label{prec-recall}%
\end{figure}

\begin{figure}[!htb]
\center{\includegraphics[scale=0.4]{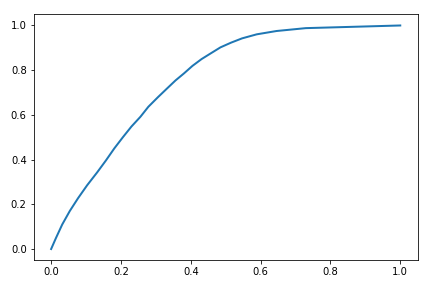}}
\caption{ROC Curve: True Positive rate vs False Positive rate}
\label{roc}
\end{figure}
To evaluate the model’s interpretability, we presented a random RBC image from the test set to the LIME explainer. We implemented our workflow in a way so that the feature computation is integrated with the prediction model. Observe that while our feature space is smaller than the input space (4 features from a $142*148$ color image), our model still captures the necessary pixels from the RBC image. 

\begin{figure*}%
\centering
\includegraphics[height=2in]{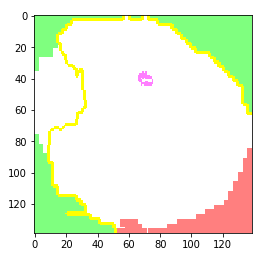}%
\caption{Demarcation of pixels that influence proposed method's prediction}
	\label{influence}
\end{figure*}

\begin{figure*}%
\centering
\includegraphics[height=2in]{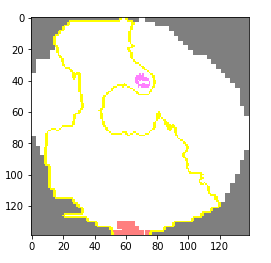}%
\caption{Demarcation of pixels that significantly influence proposed method's prediction}
\label{significant-influence}
\end{figure*}

In \ref{influence}, the green area shows the pixels that influence proposed method's prediction to be `uninfected' and the red area shows the pixels
that influence the prediction to be `infected'. The yellow boundary shows the most important pixels in the decision making procedure. Observe
that the area outside of the RBC image also influences the prediction. However, the influence is insignificant. Figure 11 shows the pixels of the image that are most significant (with weight 0.08 or larger) for proposed method's prediction. The range of weight of a pixel varies from $0$ to $1$, where weight $1$ means that changing the intensity of that single pixel can reverse the prediction. From \ref{significant-influence}, we see the most significant pixels come from the ring entity of the infected RBC image. This justifies the validity of our classifier in diagnostic decision. 

Notice that our feature computation workflow is \emph{completely automated} compared to \cite{varma2019detection} and the feature computation step is not complete black-box \cite{rajaraman2018pre}.

\section{Remarks and future work}
While several approaches have tried to use ML in predicting from RBC images, most of these methods rely on the ML model feature extractors to generate a large number of (possibly not very relevant) features. Using an increased feature-space biases the ML model and results in the so called ``curse-of-dimensionality'' problem. Our main contribution is to develop a small subset of aggregate features that result in a more compact and interpretable model while maintaining a high level of accuracy.
Our work has some limitations that present several opportunities for future work. First, the model itself can be improved in various ways. For example, we use a simple random forest model for detection. More sophisticated models can substantially improve the detection accuracy of infected images. Personalized models that use other physiological data of each patient to calibrate the model may provide even better performance. Second, detection of other related infected images in medical analysis using our modeling approach seems important future work. Third, our model was built on preprocessed data. Hence, it may not work for detecting infected RBC with severe noise in it. Finally, application of our method to other datasets can further improve its interpretability and accuracy.

\section{Conclusion}
We presented CIDMP, a system that uses reduced feature space to predict malaria-infected RBC. We performed statistical analysis on feature space to check the validity of the features in predictions.
We extracted fully explainable mathematically sound features to train the machine learning model. We employed an interpretable Random Forest Model that predicts the infected RBC with good results. Through experimentation on real-life RBC image datasets, we demonstrated that our algorithm and explainer can be used in real-life scenarios. This work opens the doors for follow up research and real-life deployment of machine learning algorithms in sensitive fields such as medical image analysis.

\bibliographystyle{plain}
\bibliography{main.bib}

\end{document}